\pdfoutput=1

\documentclass[11pt]{article}

\usepackage{EMNLP2023}

\usepackage{times}
\usepackage{latexsym}
\usepackage{amsmath}
\usepackage{amsmath,algorithm,float,tabularx,amsfonts}
\usepackage{algpseudocode}
\makeatletter
\newcommand{\multiline}[1]{%
    \begin{tabularx}{\dimexpr0.9\linewidth-\ALG@thistlm}[t]{@{}X@{}}
        #1
    \end{tabularx}
}

\newcommand{\Step}[1]{\algrenewcommand{\alglinenumber}[1]{Step ##1: } #1}

\usepackage[T1]{fontenc}
\usepackage{booktabs}
\usepackage[utf8]{inputenc}

\usepackage{microtype}
\usepackage{graphicx}
\usepackage{inconsolata}

\makeatletter
\newcommand*{\rom}[1]{\expandafter\@slowromancap\romannumeral #1@}
\makeatother

\title{DomainInv: Domain Invariant Fine Tuning and Adversarial Label Correction For QA Domain Adaptation}

\author{Anant Khandelwal \\
  Applied Scientist, Amazon
  }

\begin{document}
\maketitle
\begin{abstract}
Existing Question Answering (QA) systems limited by the capability of answering questions from unseen domain or any out-of-domain distributions making them less reliable for deployment to real scenarios. Most importantly all the existing QA domain adaptation methods are either based on generating synthetic data or pseudo labeling the target domain data. The domain adaptation methods based on synthetic data and pseudo labeling suffers either from the requirement of computational resources or an extra overhead of carefully selecting the confidence threshold to separate the noisy examples from being in the training dataset. In this paper, we propose the unsupervised domain adaptation for unlabeled target domain by transferring the target representation near to source domain while still using the supervision from source domain. Towards that we proposed the idea of domain invariant fine tuning along with adversarial label correction to identify the target instances which lie far apart from the source domain, so that the feature encoder can be learnt to minimize the distance between such target instances and source instances class wisely, removing the possibility of learning the features of target domain which are still near to source support but are ambiguous. Evaluation of our QA domain adaptation method namely, \texttt{DomainInv}\footnote{Code will be released upon acceptance} on multiple target QA dataset reveal the performance improvement over the strongest baseline.
\end{abstract}

\section{Introduction}
Over the past few years, machine learning models are widely deployed in the production but however to make them work satisfactorily in production require substantial amount of high quality annotated data, which is expensive and time consuming. Therefore it is of utmost importance to build generalizable models to be able to perform well on unseen datasets. However, due to the mechanism of \textit{domain shift} or \textit{bias} in training dataset \cite{36364, NIPS2006_b1b0432c}, it is challenging to directly transfer knowledge from the model trained on source domain to unlabeled target domain. In this paper we studied this phenomenon specifically for the case of extractive Question Answering (QA) systems. 

Extractive QA systems performs the task of finding most relevant answer for a given question in a context or paragraph. The answer is represented as the sub span of the context with start and end positions predicted by the QA model. The training data for QA essentially consists of triplets specifying the question, answer and context. The input to the model is question and context as a running text separated by a separator. The model is trained to predict the most relevant start and end positions in the context \cite{DBLP:journals/corr/SeoKFH16, chen-etal-2017-reading, devlin-etal-2019-bert, kratzwald-etal-2019-rankqa}. These QA systems also suffer from performance degradation at test time, since the question and context can be of any type, it might be the simplest and complex ways of asking the same thing, while the answer may be the reasoning based or follow any complex extraction pattern which is difficult to generalize to all the cases with limited annotated training dataset. This has been studied in some of the recent works \cite{fisch-etal-2019-mrqa, 10.5555/3524938.3525579, zengattacking} with a work around solutions of using the labeled target domain data or feedback during training\cite{daume-iii-2007-frustratingly, kratzwald-etal-2020-learning, kamath-etal-2020-selective}, while some \cite{yue2022qa, yue-etal-2021-contrastive} uses synthetic data or pseudo labeled data to train such systems to make them generalize to out-of-domain distributions. However, the pseudo labeled data is prone to noise and obtaining correctly labeled data requires lot of human labelling effort. 

\begin{figure*}
    \centering
    \includegraphics[width=\textwidth]{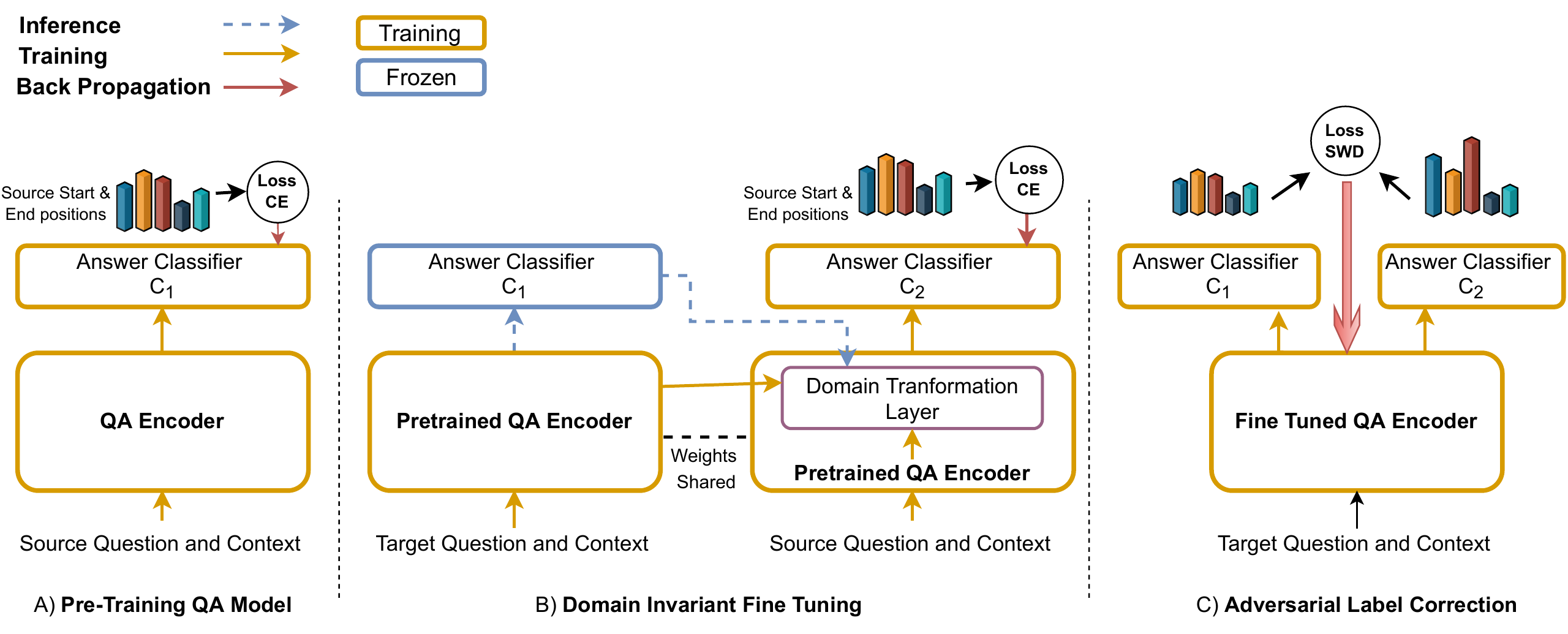}
    \caption{Proposed QA domain adaptation framework namely, \texttt{DomainInv}. It proposed to use the domain invariant fine tuning followed by adversarial label correction to mitigate the limitation of doamin invariant fine tuning.}
    \label{fig:overview}
\vspace{-4mm}
\end{figure*}

In this paper we focus on unsupervised domain adaptation (UDA) which does not require labeled target domain data. There are lot works towards achieving domain invariant representations, categorising into 1) optimizing the discrepancy between domain representations \cite{yue2022qa, yue-etal-2021-contrastive}, and 2) adversarial learning \cite{lee-etal-2019-domain, cao2020unsupervised}. For minimizing domain discrepancy existing works leveraged maximum mean discrepancy (MMD) \cite{NIPS2006_e9fb2eda} as the distance measure the distance between the source and target domain distributions. Similar to MMD, CMD (central moment discrepancy)\cite{DBLP:conf/iclr/ZellingerGLNS17}, Wasserstein distance (WD) \cite{10.5555/3504035.3504532}, sliced wasserstein distance (SWD) \cite{kolouri2019generalized}, multi-kernel MMD \cite{10.5555/3045118.3045130}, joint MMD \cite{10.5555/3305890.3305909} are other alternate measures. Inspired by generative adversarial networks (GAN)\cite{NIPS2014_5ca3e9b1}, the adaptation methods based on adversarial learning have also shown promising results \cite{Ganin2017, pmlr-v80-xie18c, 10.5555/3504035.3504517, 8578490, sliced-wasserstein-discrepancy-for-unsupervised-domain-adaptation}. Adversarial learning methods propose the idea of using the domain discriminator to distinguish the incoming sample is either from source domain and target domain, while the feature generator tries to fool the discriminator by generating domain invariant features. In this process of generating domain invariant representations, the generator generates the target representation near to source domain decision boundaries but are however are misaligned with respect to the source classes leading to performance degradation \cite{sliced-wasserstein-discrepancy-for-unsupervised-domain-adaptation}. Some works rely on high confidence pseudo labels \cite{yue2022qa, 9008112} for target domain. However, this way of generating synthetic data for target domain is an extra computation overhead. Moreover, target pseudo labeling can have adverse effects on adaptation if they generates too many incorrect labels above the confidence threshold, some works propose to minimize the distance between tokens from the target instances and that of source support contrastively \cite{yue2022qa}. But however in the practical scenario the target domain can be completely asymmetrical and hence requires the need to align the pre-trained source model with target domain before optimizing for domain invariant representations.\\
In this paper we proposed a adaptation framework called \texttt{\textbf{DomainInv}} (shown in Figure \ref{fig:overview}), which can perform the task of domain adaptation without training of answer classifier with noisy pseudo-labeled data and hence eliminate the effort to filter out that noise before training on target domain as opposed to what is required in \cite{yue2022qa}. We propose the approach of learning domain invariant features using domain invariant fine tuning along with adversarial label correction to identify such target instances which are far apart from the source domain and optimize them to lie near the source support class wisely. \textbf{Main Contributions of this paper are as follows}:
\begin{itemize}
    \item We propose the unsupervised domain adaptation framework called \texttt{DomainInv} for QA. The framework can tackle the \textit{domain shift} phenomenon without the need of explicit training of answer classifier with pseudo labeled data. The noise in pseudo labeled data (which is hard to filter out) deteriorates the performance of answer classifier and hence not able to generalize well on the target domain. 
    \item We propose the idea of 1) Domain Invariant Fine Tuning and 2) Adversarial Label correction together which minimize the distance between the source and target domain representations class wisely in an iterative manner.
    \item We evaluated our framework on multiple QA datasets as target domains without accessing their answers during training. \textit{DomainInv} outperforms the strongest baseline for QA domain adaption which adapt the model by explicitly training on pseudo labeled target domain.
\end{itemize}
\vspace{-2mm}
\section{Related Work}
In past few years, there is an increasing interest towards learning generalized representations through various learning paradigms namely, unsupervised, multi-tasking, and transfer learning \cite{ peters-etal-2018-deep, DBLP:journals/corr/abs-1806-08730, chronopoulou-etal-2019-embarrassingly, DBLP:journals/corr/abs-1811-01088, wang-etal-2018-glue, xu-etal-2019-multi}. Explicitly, there are recent studies which have explored the generalization capability of  reading comprehension systems \cite{golub-etal-2017-two, fisch2019mrqa, talmor2019multiqa, yue-etal-2021-contrastive, yue2022qa, yue-etal-2022-domain}. Only the unsupervised approaches (target domain data is unlabeled) for domain adaptation, is our interest in this paper. The approaches used for unsupervised domain adaptation are broadly categorised in the following main themes: 1) Contrastive Learning 2) Self-Supervision and 3) Adversarial Learning.

\textbf{Contrastive Learning for Self-Supervision}: Contrastive learning methods \cite{9157636, 10.5555/3495724.3496555, 10.5555/3524938.3525087, yue2022qa, yue-etal-2021-contrastive} aim to learn an feature encoder that generates similar features of the same input (obtained from different augmentations) and different features for any other input and its augmentations. Specifically for QA, \cite{sun-etal-2018-answer, du-etal-2017-learning} have generated synthetic QA samples by Question Generation (QG). Leveraging these samples improves performance in out-of-domain distribution \cite{yue-etal-2021-contrastive, golub-etal-2017-two, DBLP:journals/corr/TangDQZ17, lee-etal-2020-generating, tang-etal-2018-learning, shakeri-etal-2020-end, yue-etal-2022-synthetic, https://doi.org/10.48550/arxiv.2210.03250}. Additionally, contrastive learning has been applied to minimize the discrepancy between source and target domain using Maximum Mean Discrepancy(MMD) \cite{NIPS2006_e9fb2eda}, they learnt to minimize the distance for averaged token features\cite{yue2022qa} among answer and non-answer tokens in source and target domains and maximize between them.

\textbf{Domain Adaptation with Self-Supervision}: There are lot of works in vision which have explored the use of self-supervision for unsupervised domain adaptation. They all are aligned to the common objective of minimizing the discrepancy(distance) between domains \cite{8954037, DBLP:journals/corr/abs-2106-05528, DBLP:journals/corr/abs-2103-15566}. However, this is the objective which is used in contrastive learning as well, but still the model learnt with contrastive learning performs better \cite{pmlr-v162-shen22d}. Apart from MMD \cite{NIPS2006_e9fb2eda} criterion used in \cite{yue2022qa}, other metrics central moment discrepancy (CMD) used in \cite{DBLP:conf/iclr/ZellingerGLNS17} to directly match order-wise differences of central moments. Wasserstein distance, which is used to measure the distance between two probability distributions, has been explored in \cite{10.5555/3504035.3504532, kolouri2019generalized}. The method in \cite{yu-etal-2020-wasserstein} learn sentence representation for text matching between asymmetrical domains. We also consider the use of sliced wasserstein distance \cite{kolouri2019generalized}, rather than minimizing distance between representation for domains, this has been applied to minimize the distribution learned for start token and end token for QA domain adaptation.\\
\textbf{Adversarial Learning}: The objective for adversarial learning is also based on the idea to minimize the domain discrepancy. The main idea of domain adversarial learning is to learn the domain invariant representation via an adversarial loss between the feature generator and discriminator similar to GANs \cite{NIPS2014_5ca3e9b1}. Some works which uses domain adversarial learning are \cite{Ganin2017, 8099799, 8099501, DBLP:journals/corr/abs-2002-04869, NEURIPS2018_ab88b157, 10.5555/3504035.3504517}. Additionally, there are methods\cite{yue2022qa} which explored the use of target data along with pseudo labels to train the target classifier. Different from this we have explored the use of adversarial loss to identify and correct the mistakes in labels during target aware source fine-tuning to learn the domain invariant representation for target domain.

\section{Setup} 
Problem setup for unsupervised domain adaptation(UDA) consider the labeled source domain $\mathcal{D}_s$ and unlabeled target domain $\mathcal{D}_t$. The goal is to maximize the performance on target domain by only training with labeled source domain data and unlabeled target domain data as in \cite{cao2020unsupervised, shakeri-etal-2020-end, yue-etal-2021-contrastive, yue-etal-2022-domain, yue2022qa}.\\
\textbf{Data}: Specifically, for the case of QA domain adaptation we describe the labeled source domain $\mathcal{D}_s$ data as samples consisting of triplets, $\{c_s^{(i)}, q_s^{(i)}, a_s^{(i)}\} \in \mathbf{X}_s$, consisting of context $c_s^{(i)}$, question $q_s^{(i)}$ and answer $a_s^{(i)}$, where each triplet is obtained from the training data $\mathbf{X}_s$. Similarly, the unlabeled target domain $\mathcal{D}_t$ data consists of samples with pair $\{c_t^{(i)}, q_t^{(i)}\} \in \mathbf{X}_t$ consisting of only context $c_t^{(i)}$ and question $q_t^{(i)}$, obtained from unlabelled training data $\mathbf{X}_t$. Here, in our case of QA domain adaptation the answer is the start and end position in the context since we are working with extractive QA systems. \\
\textbf{Model}: We approach the problem of QA domain adaptation as training the model function $\mathit{f}$ which predicts an answer $a_t^{(i)}$ given the context $c_t^{(i)}$ and question $q_t^{(i)}$ from $\mathbf{X}_t$, denoted as $a_t^{(i)} = \mathit{f}(c_t^{(i)}, q_t^{(i)})$. This requires to optimize the function $\mathit{f}$ for maximum performance on target domain $\mathcal{D}_t$. Mathematically, this is denoted as:
\begin{equation}
\underset{{\mathit{f}}}{min} \textrm {  } \mathcal{L}(\mathit{f}, \mathbf{X}_t)
\end{equation}
where $\mathcal{L}$ is the loss function. We adopt the two fold training scheme to maximize performance on target domain namely, Domain Invariant Fine Tuning and Adversarial Label Correction which will be discussed in the following sections.

\section{The \texttt{DomainInv} Framework}
\subsection{Overview}
The proposed \texttt{DomainInv} framework consists of two main components: 1) \textbf{Domain Invariant Fine Tuning} 2) \textbf{Adversarial Label Correction} for domain adaptation as shown in Figure \ref{fig:overview}. We consider having the pre-trained QA model $\mathit{f}$, fine-tuned on the source domain $\mathcal{D}_s$ as in \cite{cao2020unsupervised} with an additional batch norm layer. The answer classifier $\mathcal{C}_1$ obtained is the one predicting the start and end index in the context. During domain invariant fine-tuning, we consider the use of target domain $\mathcal{D}_t$, to augment the style of pseudo answer and non-answer tokens to source domain, this results in another answer classifier $\mathcal{C}_2$, which has the target domain style information while still being trained on the source domain. With answer classifier $\mathcal{C}_2$ there are some instances in target domain $\mathcal{D}_t$ for which the answer is not same as the one obtained using $\mathcal{C}_1$. We consider these instances as the one which are far apart from the source domain and the QA model is least confident about these so we identify such instances during adversarial correction. This is due to answer classifier $\mathcal{C}_2$ contains the style information of target domain, and $\mathcal{C}_1$ contains the only information of source domain. During adversarial label correction, we minimize the distribution between these two and update the BERT encoder to generate the features for the target domain closer to the source domain, so that the classifier $\mathcal{C}_2$ predict the answer as if it was doing for source domain. 
\subsection{Domain Invariant Fine Tuning}
\begin{figure}
    \centering
    \includegraphics[scale=0.5]{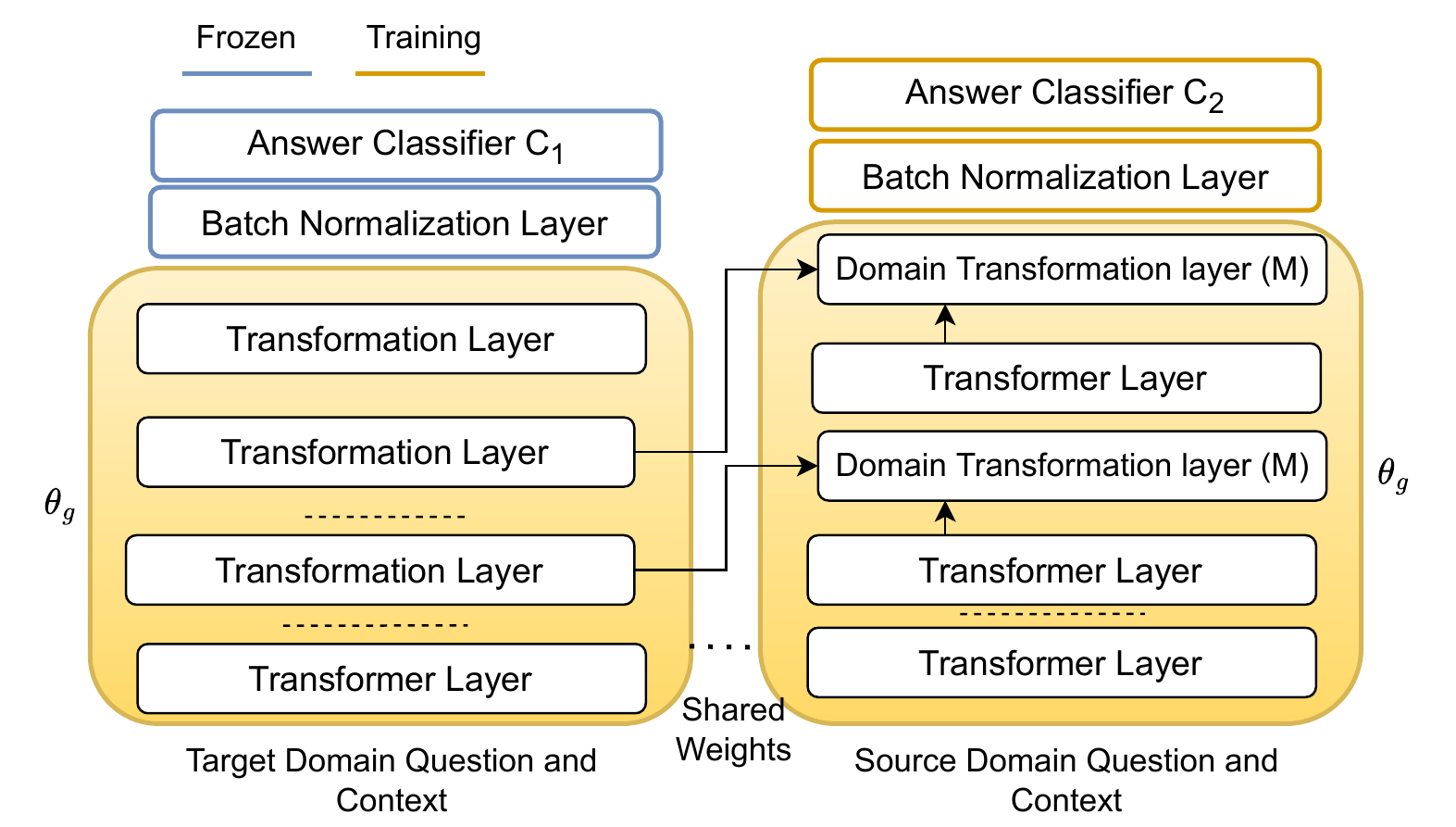}
    \caption{Domain Invariant Fine Tuning with Domain Transformation Layer}
    \label{fig:layer}
\vspace{-4mm}
\end{figure}

In this section we will explain in detail the process we have followed for domain invariant fine tuning. Consider the fine-tuned QA model $\mathit{f}$ is a BERT model with $\mathbf{L}$ layers of transformers\cite{vaswani2017attention}. Specifically, let $\mathcal{C}_1$ be an answer classifier and $\theta_g$ be the encoder parameters for fine-tuned QA model. During domain invariant fine tuning (shown in Figure \ref{fig:layer}) we propose to feed the style information of target domain $\mathcal{D}_t$ to the source domain at each layer $l \in \mathbf{L}$ as shown in Figure \ref{fig:layer}. We keep the weights shared between the two encoders to allow the target domain information be updated in the BERT encoder with the supervision of source domain. Let $\phi(x, x')$ be the learnable domain shift vector between the source instance $x$ and target domain instance $x'$, and $\mathcal{M}(x, \phi(x, x'))$ be a learnable domain transformation layer which when introduced at the top of each transformer layer $l \in \mathbf{L}$ of model $f$, transforms the parameters of source domain classifier $\mathcal{C}_1$ to target aware classifier $\mathcal{C}_2$ and update encoder parameters $\theta_g$ with style of target domain.\\
\textbf{Domain Transformation Layer $\mathcal{M}$}: The domain transformation layer is expected to fuse the domain shift vector with the hidden states (which were fine tuned for source domain) at each layer of transformer. The domain shift vector $\phi$ should solely captures the information which are different from the source domain. This categorizes the vector which contains any extra information in target domain as compared to the source domain irrespective of its position in the context. This can be achieved by taking the difference between the average pooled vector of hidden states at each layer obtained for source and target domain. Lets say the $\mathbf{H}_s^{(l)}$ and $\mathbf{H}_t^{(l)}$ be the hidden states obtained at layer $l$ for the source domain and target domain respectively. Then the domain shift vector at layer $l$ between two instances is given as:
\begin{equation}
    \phi^{(l)}(H_t^{(l)}, H_s^{(l)})  = avg(W H_t^{(l)}) - avg(W H_s^{(l)})
\end{equation}
where, $W \in \mathbf{R}^{k \times d}$ are the linear transform parameters shared across layers. Then, the domain transformation layer $\mathcal{M}$ is given as:
\begin{equation}
\begin{aligned}
    \mathcal{M}(H_s^{(l)}, \phi^{(l)}(H_t^{(l)}, H_s^{(l)})) = H_s^{(l)} + \\ W^T \phi^{(l)}(H_t^{(l)}, H_s^{(l)})
\end{aligned} 
\end{equation}
where $W^T \phi^{(l)}(H_t^{(l)}, H_s^{(l)}) \in \mathbf{R}^d$ is added to all hidden states at layer $l$ corresponding to source domain. This design of domain shift vector follows the identity property i.e. $\phi(x, x) = 0$, this allows us to plug in the domain transformation layer only at the time of training while at the time of inference for target domain, the $\phi^{(l)}(H_t^{(l)}, H_t^{(l)})) = 0$. Still it is required to carefully choose the value of $k$, which is a hyper-parameter, because it assumes the domain shift information lies in the $k$-dimensional subspace where the difference between two domains can be minimized to make them appear similar. \\
\textbf{QA Domain Transformation}: Specifically, for the case of QA domain adaptation we cant apply the domain shift across all the tokens invariably, since there are underlying difference between the context, question and answer tokens. Hence, it would be wise to apply to calculate $\phi_A$, $\phi_C$ and $\phi_Q$ for Answer, Context and Questions respectively. Since we only have context and questions in the target domain, the answer is the pseudo answer obtained from the classifier $\mathcal{C}_1$, where the weights of $\mathcal{C}_1$ and frozen and the BERT encoder parameters $\theta_g$ are shared during fine-tuning, those will be updated jointly along with classifier $\mathcal{C}_2$ (initialized with fine tuned classifier $\mathcal{C}_1$) as shown in Figure \ref{fig:layer}. At the end of domain transformation fine-tuning, we get another target aware classifier $\mathcal{C}_2$ and updated encoder parameters $\theta_g$ which are the fine-tuned parameters on target aware source domain with supervised cross-entropy loss $\mathcal{L}_{ce}$ i.e. 
\begin{equation}
\underset{{\mathit{f}}}{min} \textrm {  } \mathcal{L}_{ce}(\mathit{f}, \mathbf{X}_s ||  \mathbf{X}_t)  
\label{eqn:finetune}
\end{equation}
where $f$ consists of parameters $\theta_g$ and parameters in classifier $\mathcal{C}_2$ and $\mathcal{M}$. During training, we pick randomly the samples of target domain of same batch size as source domain with an additional constraint to contain the parallel instances having the same question types (denoted as $\mathbf{X}_s ||  \mathbf{X}_t$). We use the dependency parser, semantic role labelling and NER to detect the question types as in \cite{keklik2018automatic}.

\subsection{Adversarial Label Correction}
We have introduced the adversarial label correction basis the matter of the fact that during domain invariant fine tuning we have relied on the pseudo labels obtained from classifier $\mathcal{C}_1$, however these labels are noisy and this is prone to error accumulation and can lead to performance degradation if there are lot of instances where the pseudo labels are incorrectly annotated. There are several works for QA domain adaptation which relies on data augmentation and pseudo labels\cite{yue2022qa, yue-etal-2021-contrastive}, but however they missed the fact that domain adaptation can result the target features near the source decision boundaries but are ambiguous and hence error prone. To mitigate this we propose the idea of reducing the inconsistency between domains by generating the target features near to the support of source classes, in our case this is the starting and ending indices for the answer classifier for different question types.

In Equation \ref{eqn:finetune}, the fine tuning happens with target aware source domain which results in answer classifier $\mathcal{C}_2$. This learns the distribution of start and end classes given the style of target domain $\mathcal{D}_t$, but these can be error prone, mainly due to two reasons: 1) The target pseudo labels obtained from $\mathcal{C}_1$ is erroneous and 2) The target domain information makes the classifier $\mathcal{C}_2$ difficult to learn the correct distribution of start and end classes on the source domain. Collectively, this happens when the target instances are far apart from the source domain and require to do explicit optimization for such cases. Hence, we make use of the adversarial loss to first identify the samples of target domain which are far apart from the support of source domain and hence we update the parameters of $\mathcal{C}_1$ and $\mathcal{C}_2$ with $\theta_g$ (obtained after domain invariant fine tuning) frozen to maximize the discrepancy due to such instances and then minimize the parameters of $\theta_g$ to generate the target domain features near to the source domain for start and end classes. Mathematically, this has been written as the minmax game of learning domain invariant representations:
\begin{equation}
    \underset{\theta_g}{\textrm{min}}\underset{\mathcal{C}_1, \mathcal{C}_2}{\textrm{ max}} \textrm{  } \mathcal{L}_{lc}(X_t)
\end{equation}
where $\mathcal{L}_{lc}$ is the label correction loss optimized for the target domain for maximum performance. The discrepancy maximizing term has been mathematically formulated as the Sliced Wasserstein Distance(SWD) between the representation as in \cite{kolouri2019generalized} learnt for start and end classes by classifier $\mathcal{C}_1$ and $\mathcal{C}_2$ respectively. Let $\mathcal{G}$ denote the BERT encoder with parameters $\theta_g$, the loss is written as:
\begin{equation}
\begin{aligned}
    \underset{{\mathit{\mathcal{C}_1, \mathcal{C}_2}}}{\textrm{min}} \textrm {  } \mathcal{L}_{ce}(\mathit{f}, \mathbf{X}_s ||  \mathbf{X}_t)  - \\ 
    \sum_{k\in \{s, e\}}\mathcal{L}_{swd}(\mathcal{C}_1^k(G(X_t)), \mathcal{C}_2^k(G(X_t)))
\end{aligned}
\label{eqn:finetune2}
\end{equation}
where $\mathcal{C}_1^k(.)$, $\mathcal{C}_2^k(.)$ for all $k \in \{s,e\}$ denotes the probability distribution obtained from classifier $\mathcal{C}_1$ and $\mathcal{C}_2$ for starting and ending indices $s$ and $e$. This loss updates both classifiers on target aware source domain only for those instances where parallel target domain instances are inconsistent. Since both $\mathcal{C}_1$ and $\mathcal{C}_2$ trained to predict the start and end indices for the source domain, one with source domain features only and another on source augmented with target domain features. Hence, the classifier $\mathcal{C}_2$ can predict the different start and end indices for those instances where $\mathcal{C}_1$ is incorrect for target domain and hence we need to update the parameters $\theta_g$ so as to generate the target domain features near to source support class wisely. This has been achieved by minimizing the loss function:
\begin{equation}
    \underset{\mathcal{G}}{\textrm{min}} \sum_{k\in \{s, e\}}\mathcal{L}_{swd}(\mathcal{C}_1^k(G(X_t)), \mathcal{C}_2^k(G(X_t)))
    \label{eqn:theta_adv}
\end{equation}
Finally, at the end we use the Encoder $\mathcal{G}$ and answer classifier $\mathcal{C}_1$ as the domain adapted QA model. End-to-end training of \texttt{DomainInv} Framework is shown in Algorithm \ref{lab:ide_algo}.
\begin{algorithm}
       \caption{\texttt{DomainInv} Training for UDA}
       \small
       \begin{algorithmic}
           \Require Labeled Source $\{\mathcal{X}_s, \mathcal{Y}_s\}$; Unlabeled Target $\{\mathcal{X}_t\}$, hyperparameter $k$, fine tuned QA model with encoder $\mathcal{G}$ and Classifier $\mathcal{C}_1$ and classifier $\mathcal{C}_2$ initialized with $\mathcal{C}_1$. \\\vspace{1mm}
            \Step \multiline{%
            \textbf{Step 1}: Update $\mathcal{G}$, $\mathcal{C}_2$ on Source Domain (with target style augmentation) using Domain Invariant Fine-Tuning as in Equation \ref{eqn:finetune}}\\
           
           \While{ $\mathcal{G}$, $\mathcal{C}_1$, $\mathcal{C}_2$ still converging} 
            
            \Step \multiline{%
            \textbf{Step 2}: Update $\mathcal{C}_1$, $\mathcal{C}_2$ on target aware source set to maximize the sliced Wasserstein distance (SWD) on target instances as in Equation \ref{eqn:finetune2}}
            
            \Step \multiline{%
            \textbf{Step 3}: Update $\mathcal{G}$  to minimize the SWD as calculated earlier according to Equation \ref{eqn:theta_adv}}
            
           \EndWhile
       \end{algorithmic}
       \label{lab:ide_algo}
\end{algorithm}

\begin{table*}[]
\begin{tabular}{llllll}
\toprule
\textbf{Model}     & \multicolumn{1}{c}{\textbf{HotpotQA}} & \multicolumn{1}{c}{\textbf{NaturalQ.}} & \multicolumn{1}{c}{\textbf{NewsQA}} & \multicolumn{1}{c}{\textbf{SearchQA}} & \multicolumn{1}{c}{\textbf{TriviaQA}} \\
                   & \multicolumn{1}{c}{EM / F1}           & \multicolumn{1}{c}{EM / F1}            & \multicolumn{1}{c}{EM / F1}         & \multicolumn{1}{c}{EM / F1}           & \multicolumn{1}{c}{EM / F1}           \\ \midrule
\multicolumn{6}{c}{(1) \textbf{Zero Shot Target Performance}}                                                                                                                                                                      \\ \midrule
BERT               & 43.34/60.42                           & 39.06/53.7                             & 39.17/56.14                         & 16.19/25.03                           & 49.70/59.09                           \\ \midrule
\multicolumn{6}{c}{(2) \textbf{QA Domain Adaptation Target Performance}}                                                                                                                                                           \\ \midrule
DAT \cite{lee-etal-2019-domain}                & 44.25/61.10                           & 44.94/58.91                            & 38.73/54.24                         & 22.31/31.64                           & 49.94/59.82                           \\
CASe \cite{cao2020unsupervised}              & 47.16/63.88                           & 46.53/60.19                            & 43.43/59.67                         & 26.07/35.16                           & 54.74/63.61                           \\
CAQA \cite{yue-etal-2021-contrastive}             & 46.37/61.57                           & 48.55/62.60                            & 40.55/55.90                         & 36.05/42.94                           & 55.17/63.23                           \\
CAQA* \cite{yue-etal-2021-contrastive, yue2022qa}             & 48.52/64.76                           & 47.37/60.52                            & 44.26/60.83                         & 32.05/41.07                           & 54.30/62.98                           \\
QADA \cite{yue2022qa}              & 50.80/65.75                           & 52.13/65.00                            & 45.64/61.84                         & 40.47/48.76                           & 56.92/65.86                           \\
DomainInv(Ours)    &     \textbf{52.92/66.71}                                  &    \textbf{54.97/68.80}                           &        \textbf{45.96/61.88}                             &          \textbf{40.92/49.88}                             &     \textbf{57.78/66.64}                                 \\ \midrule
\multicolumn{6}{c}{(3) \textbf{Supervised Training Target Performance}}                                                                                                                                                            \\ \midrule
BERT (10K Samples) &      49.57/66.65                               &       54.81/67.98                                  &       45.92/61.85                              &        60.21/66.96                               &                     53.87/60.42                  \\
BERT (All Samples) & 57.96/74.76                           & 67.08/79.02                            & 52.14/67.46                         & 71.54/77.77                           & 64.51/70.27                           \\ \bottomrule
\end{tabular}
\caption{QA Adaptation Performance on Target Domains}
\label{tab:results}
\vspace{-4mm}
\end{table*}

\section{Experiments}
\subsection{Experimental Setup}
\label{lab:exp_setup}
\textbf{Datasets}: We consider the source domain $\mathcal{D}_s$ as \textbf{SQuAD v1.1} \cite{rajpurkar-etal-2016-squad} following \cite{yue2022qa, yue-etal-2021-contrastive, shakeri-etal-2020-end, cao2020unsupervised, lee-etal-2020-generating}. \textbf{SQuAD v1.1} is a well known annotated QA dataset where paragraphs (context) are from Wikipedia articles. Target Domains $\mathcal{D}_t$ are considered from MRQA Split \rom{1} \cite{fisch2019mrqa}, namely, \textbf{NaturalQuestions} \cite{10.1162/tacl_a_00276}, \textbf{HotpotQA} \cite{yang-etal-2018-hotpotqa}, \textbf{SearchQA} \cite{DBLP:journals/corr/DunnSHGCC17}, \textbf{TriviaQA} \cite{joshi-etal-2017-triviaqa}, \textbf{NewsQA} \cite{https://doi.org/10.48550/arxiv.1611.09830}. The description of each dataset is given in Appendix \ref{sec:appendix}

\textbf{Baselines}: We trained our \texttt{DomainInv} Framework on top of the fine-tuned QA model adopted from BERT with additional batch normalization layer as in \cite{cao2020unsupervised}. This fine tuned BERT model which is trained on source domain act as the naive baseline. However, to further consider the robustness of our framework in QA domain adaptation, we adopt the following SOTA (state-of-the-art) baselines: 
1) \textbf{QADA} \cite{yue2022qa}: QA Domain Adaptation(QADA) leverages hidden space augmentation for enriching training dataset and attention based contrastive learning 2) \textbf{CAQA} \cite{yue-etal-2021-contrastive}: Contrastive Domain Adaption for Question Answering (CAQA) combines question generation and contrastive domain adaptation to learn domain-invariant features, so that it can capture both domains and thus transfer knowledge to the target distribution 3) \textbf{DAT} \cite{8099799, lee-etal-2019-domain}: Domain Adversarial Training (DAT) follows the known adversarial training and it uses the [CLS] token in BERT as discriminator to learn the generalized features from both source and target domain after training with labeled source domain 4) \textbf{CAQA*} \cite{yue-etal-2021-contrastive, yue2022qa}:  Instead of question generation in CAQA this baseline used the same process of generating pseudo labels and self-supervised adaptation as in QADA. 5) \textbf{CASe} \cite{cao2020unsupervised}: Conditional adversarial self-training(CASe) is an unsupervised domain adaptation method which iteratively do self-training on high-confidence pseudo labels  and conditional adversarial learning.\\
\textbf{Training, Evaluation and Implementation}: Following \cite{cao2020unsupervised, yue2022qa} we train the naive baseline of BERT model with an additional batch norm layer after encoder (in PyTorch by Hugging Face, using the base-uncased pretrained model with 12 layers and 768-dim hidden state). Specifically, we used learning rate of $3 \cdot 10^{-5}$ and training for $2$ epochs with batch size of $12$ and optimized using AdamW optimizer with 10\% linear warmup on source domain $\mathcal{D}_s$. Following \cite{lee-etal-2020-generating, shakeri-etal-2020-end, yue-etal-2021-contrastive} we evaluate exact matches (EM) and F1 score on the dev sets. Rest of the baselines are implemented as per the methods described in their corresponding paper. For \texttt{\textbf{DomainInv}} Framework, we ran the domain invariant fine tuning followed by adversarial label correction and repeat this for $10$ epochs with the AdamW optimizer, learning rate of $10^{-5}$, with $10\%$ linear warmup. During fine tuning we generate the labels for the target domain using the classifier $\mathcal{C}_1$ which is frozen during fine tuning and have sampled parallel samples of target domain with question types of source domain samples in a given batch size of 12. During fine-tuning there is only one hyperparameter named $k$ for domain transformation layer which has been searched for best value in [64, 128, 256, 512, 768]. Eventually the best value of $256$ works for us in almost all the cases and is the one with the maximum performance on source domain $\mathcal{D}_s$ during fine tuning.  After domain invariant fine tuning the obtained classifier $\mathcal{C}_1$, $\mathcal{C}_2$ and encoder $\mathcal{G}$ is trained with adversarial label correction. We stopped the training in between if there is no decrease in loss described in equation \ref{eqn:theta_adv} for the continuous $3$ epochs in a row.

\begin{table*}[ht]
\begin{tabular}{llllll}
\toprule
\textbf{Model}                   & \multicolumn{1}{c}{\textbf{HotpotQA}} & \multicolumn{1}{c}{\textbf{NaturalQ.}} & \multicolumn{1}{c}{\textbf{NewsQA}} & \multicolumn{1}{c}{\textbf{SearchQA}} & \multicolumn{1}{c}{\textbf{TriviaQA}} \\
                                 & \multicolumn{1}{c}{EM / F1}           & \multicolumn{1}{c}{EM / F1}            & \multicolumn{1}{c}{EM / F1}         & \multicolumn{1}{c}{EM / F1}           & \multicolumn{1}{c}{EM / F1}           \\ \midrule
DomainInv(Ours)    &     \textbf{52.92/66.71}                                  &    \textbf{54.97/68.80}                           &        \textbf{45.96/61.88}                             &          \textbf{40.92/49.88}                             &     \textbf{57.78/66.64}                                   \\
w/o Adversarial Label Correction &   51.60/64.07                                    &      53.91/65.21                                  &     45.88/61.86                                &     39.81/46.98                                  &           56.98/65.32                            \\ \bottomrule
\end{tabular}
\caption{Ablation Study: QA Adaptation Performance on Target Domains by different components of \texttt{DomainInv}}
\label{tab:tab_2}
\vspace{-4mm}
\end{table*}
\subsection{Experimental Results}
Table \ref{tab:results} shows the results for QA domain adaption performance on various target domains as described in Section \ref{lab:exp_setup}. We grouped our results and analysis into the three main categories namely 1) \textbf{Zero short Target Performance}: This is to report the results on target domain with BERT fine-tuned model without any domain adaptation on target domain this acts as a \textit{lower bound} for the domain adaptation approaches. 2) \textbf{QA Domain Adaptation Target Performance}, this one report the results due to various domain adaptation methods including \texttt{DomainInv}. 3) \textbf{Supervised Training Target Performance}: this report the result after supervised training of BERT on the target domain with randomly chosen 10K samples as well as all samples to set the \textit{upper bound} performance of QA domain adaptation approaches. 
QA domain adaptation performance (shown in Table \ref{tab:results}) using \texttt{DomainInv} Framework outperforms all the domain adaptation baselines across all the target domains and is well beyond the naive baseline. In fact almost all the domain adaptation baselines outperformed the naive baseline by a significant margin on all the target domains. However BERT fails badly (as compared to domain adaptation baselines) on some of the target domains namely, Natural Questions and SearchQA due to two main reasons: 1) BERT does not understand the style of Natural Questions even if the Wikipedia article is same the real user questions style is different than the one asked in SQuAD v1.1.  2) BERT does not understand the long form of contexts which is usual in SearchQA and it learns to focus the nearby of similar tokens as appeared in SQuAD v1.1 but the actual or more effective answer is also present in the long context. As compared to the worst QA adaptation baseline it outperforms the BERT on these two domains on average by $2.64\%$ in EM and the other domains by $1.2\%$ in EM. This is the main reason we adopted the domain style based transformation layer and the corresponding fine tuning which can make the BERT understand the different contexts as well as the questions. \texttt{DomainInv} Framework outperforms all, on average it outperforms the best baseline by $2.59\%$ and $2.17\%$ in EM and F1 respectively. Moreover, our framework outperforms the supervised training with 10K target data additionally on Natural Questions and NewsQA as compared to QADA (best baseline) which outperforms supervised baseline of 10K target data only on HotpotQA and TriviaQA.
\vspace{-2mm}
\subsection{Ablation Studies}
In Table \ref{tab:results} we compared \texttt{DomainInv} framework against the strongest baseline named QADA. However, this comparison does not detail out the importance of each component of \texttt{DomainInv} namely, domain invariant fine tuning and adversarial label correction, absence of these can cause the maximum drop of $5.17\%$ as compared to the best baseline QADA. This clears the advantage of using \texttt{DomainInv} over other domain adaptation approaches. However, the contribution of each component towards performance gain is still unknown. Hence, we studied the performance (mentioned in Table \ref{tab:tab_2}) of \texttt{DomainInv} after removing the adversarial label correction component only since removing the first component will leave only the BERT pre-trained model for which the zero shot performance is already mentioned in the Table \ref{tab:results}.  The performance drop in Table \ref{tab:tab_2} depicts clearly the advantage of adversarial label correction. For target domains namely, HotpotQA, Natural Questions and NewsQA, the performance of \texttt{DomainInv w/o adversarial label correction} in terms of EM are still higher than that of QADA, while in SearchQA it goes below the performance of QADA. This indicates the important insight on the functioning of adversarial label correction, for long contexts like in SearchQA where the answer to match exactly requires significant correction, while in the Natural Questions despite the question style has been changed but the context is still the same i.e. Wikipedia articles, hence requiring small correction in labels. This proves the effectiveness of label correction component in \texttt{DomainInv} framework.  

\section{Conclusion}
In this paper we proposed a novel QA domain adaptation framework called \texttt{DomainInv}, an unsupervised algorithm which does not require the use of labeled target domain, neither it depends on the synthetic data or pseudo labeled target domain. \texttt{DomainInv} uses 1) Domain Invariant Fine Tuning which fine tunes the QA model using the target style on the source domain and 2) Adversarial Label Correction which identifies the target distributions which are still far apart from source domain and optimize the feature generator to bring them closer near to source support class wisely. Evaluation of \texttt{DomainInv} showed that it outperforms all the baselines and achieves the superior performance establishing the new benchmark on QA domain adaptation.

\section*{Limitations}
In this section we highlight some of the limitations of \texttt{DomainInv} which has not been discussed in the paper. During domain invariant fine tuning we propose the fine tuning with new layer named domain adaptation layer which takes in the difference between average pooled source and target domain representation, but however this design choice considered all the tokens in source domain and target domain of equal importance but however this cant be the case the token distribution after each layer is also influenced by all the tokens during the self attention mechanism. Hence we, should also consider the use of attention weighted representations before taking the difference between the two. Also, during the adversarial label correction we proposed to only use the target domain to adjust the feature encoder but we should also use the source domain as well to jointly align them so that source domain representation will remain intact, future work should explore these line of thoughts.

\bibliography{anthology,custom}

\begin{thebibliography}{67}
\expandafter\ifx\csname natexlab\endcsname\relax\def\natexlab#1{#1}\fi

\bibitem[{Ben-David et~al.(2010)Ben-David, Blitzer, Crammer, Kulesza, Pereira,
  and Vaughan}]{36364}
Shai Ben-David, John Blitzer, Koby Crammer, Alex Kulesza, Fernando Pereira, and
  Jennifer Vaughan. 2010.
\newblock \href {http://www.springerlink.com/content/q6qk230685577n52/} {A
  theory of learning from different domains}.
\newblock \emph{Machine Learning}, 79:151--175.

\bibitem[{Ben-David et~al.(2006)Ben-David, Blitzer, Crammer, and
  Pereira}]{NIPS2006_b1b0432c}
Shai Ben-David, John Blitzer, Koby Crammer, and Fernando Pereira. 2006.
\newblock \href
  {https://proceedings.neurips.cc/paper/2006/file/b1b0432ceafb0ce714426e9114852ac7-Paper.pdf}
  {Analysis of representations for domain adaptation}.
\newblock In \emph{Advances in Neural Information Processing Systems},
  volume~19. MIT Press.

\bibitem[{Bousmalis et~al.(2017)Bousmalis, Silberman, Dohan, Erhan, and
  Krishnan}]{8099501}
K.~Bousmalis, N.~Silberman, D.~Dohan, D.~Erhan, and D.~Krishnan. 2017.
\newblock \href {https://doi.org/10.1109/CVPR.2017.18} {Unsupervised
  pixel-level domain adaptation with generative adversarial networks}.
\newblock In \emph{2017 IEEE Conference on Computer Vision and Pattern
  Recognition (CVPR)}, pages 95--104, Los Alamitos, CA, USA. IEEE Computer
  Society.

\bibitem[{Cao et~al.(2020)Cao, Fang, Yu, and Zhou}]{cao2020unsupervised}
Yu~Cao, Meng Fang, Baosheng Yu, and Joey~Tianyi Zhou. 2020.
\newblock Unsupervised domain adaptation on reading comprehension.
\newblock In \emph{Proceedings of the AAAI Conference on Artificial
  Intelligence}, volume~34, pages 7480--7487.

\bibitem[{Caron et~al.(2020)Caron, Misra, Mairal, Goyal, Bojanowski, and
  Joulin}]{10.5555/3495724.3496555}
Mathilde Caron, Ishan Misra, Julien Mairal, Priya Goyal, Piotr Bojanowski, and
  Armand Joulin. 2020.
\newblock Unsupervised learning of visual features by contrasting cluster
  assignments.
\newblock In \emph{Proceedings of the 34th International Conference on Neural
  Information Processing Systems}, NIPS'20, Red Hook, NY, USA. Curran
  Associates Inc.

\bibitem[{Chen et~al.(2017)Chen, Fisch, Weston, and
  Bordes}]{chen-etal-2017-reading}
Danqi Chen, Adam Fisch, Jason Weston, and Antoine Bordes. 2017.
\newblock \href {https://doi.org/10.18653/v1/P17-1171} {Reading {W}ikipedia to
  answer open-domain questions}.
\newblock In \emph{Proceedings of the 55th Annual Meeting of the Association
  for Computational Linguistics (Volume 1: Long Papers)}, pages 1870--1879,
  Vancouver, Canada. Association for Computational Linguistics.

\bibitem[{Chen et~al.(2020)Chen, Kornblith, Norouzi, and
  Hinton}]{10.5555/3524938.3525087}
Ting Chen, Simon Kornblith, Mohammad Norouzi, and Geoffrey Hinton. 2020.
\newblock A simple framework for contrastive learning of visual
  representations.
\newblock In \emph{Proceedings of the 37th International Conference on Machine
  Learning}, ICML'20. JMLR.org.

\bibitem[{Chronopoulou et~al.(2019)Chronopoulou, Baziotis, and
  Potamianos}]{chronopoulou-etal-2019-embarrassingly}
Alexandra Chronopoulou, Christos Baziotis, and Alexandros Potamianos. 2019.
\newblock \href {https://doi.org/10.18653/v1/N19-1213} {An embarrassingly
  simple approach for transfer learning from pretrained language models}.
\newblock In \emph{Proceedings of the 2019 Conference of the North {A}merican
  Chapter of the Association for Computational Linguistics: Human Language
  Technologies, Volume 1 (Long and Short Papers)}, pages 2089--2095,
  Minneapolis, Minnesota. Association for Computational Linguistics.

\bibitem[{Daum{\'e}~III(2007)}]{daume-iii-2007-frustratingly}
Hal Daum{\'e}~III. 2007.
\newblock \href {https://aclanthology.org/P07-1033} {Frustratingly easy domain
  adaptation}.
\newblock In \emph{Proceedings of the 45th Annual Meeting of the Association of
  Computational Linguistics}, pages 256--263, Prague, Czech Republic.
  Association for Computational Linguistics.

\bibitem[{Deng et~al.(2019)Deng, Luo, and Zhu}]{9008112}
Zhijie Deng, Yucen Luo, and Jun Zhu. 2019.
\newblock \href {https://doi.org/10.1109/ICCV.2019.01004} {Cluster alignment
  with a teacher for unsupervised domain adaptation}.
\newblock In \emph{2019 IEEE/CVF International Conference on Computer Vision
  (ICCV)}, pages 9943--9952.

\bibitem[{Devlin et~al.(2019)Devlin, Chang, Lee, and
  Toutanova}]{devlin-etal-2019-bert}
Jacob Devlin, Ming-Wei Chang, Kenton Lee, and Kristina Toutanova. 2019.
\newblock \href {https://doi.org/10.18653/v1/N19-1423} {{BERT}: Pre-training of
  deep bidirectional transformers for language understanding}.
\newblock In \emph{Proceedings of the 2019 Conference of the North {A}merican
  Chapter of the Association for Computational Linguistics: Human Language
  Technologies, Volume 1 (Long and Short Papers)}, pages 4171--4186,
  Minneapolis, Minnesota. Association for Computational Linguistics.

\bibitem[{Du et~al.(2017)Du, Shao, and Cardie}]{du-etal-2017-learning}
Xinya Du, Junru Shao, and Claire Cardie. 2017.
\newblock \href {https://doi.org/10.18653/v1/P17-1123} {Learning to ask: Neural
  question generation for reading comprehension}.
\newblock In \emph{Proceedings of the 55th Annual Meeting of the Association
  for Computational Linguistics (Volume 1: Long Papers)}, pages 1342--1352,
  Vancouver, Canada. Association for Computational Linguistics.

\bibitem[{Dunn et~al.(2017)Dunn, Sagun, Higgins, G{\"{u}}ney, Cirik, and
  Cho}]{DBLP:journals/corr/DunnSHGCC17}
Matthew Dunn, Levent Sagun, Mike Higgins, V.~Ugur G{\"{u}}ney, Volkan Cirik,
  and Kyunghyun Cho. 2017.
\newblock \href {http://arxiv.org/abs/1704.05179} {Searchqa: {A} new q{\&}a
  dataset augmented with context from a search engine}.
\newblock \emph{CoRR}, abs/1704.05179.

\bibitem[{Fisch et~al.(2019{\natexlab{a}})Fisch, Talmor, Jia, Seo, Choi, and
  Chen}]{fisch-etal-2019-mrqa}
Adam Fisch, Alon Talmor, Robin Jia, Minjoon Seo, Eunsol Choi, and Danqi Chen.
  2019{\natexlab{a}}.
\newblock \href {https://doi.org/10.18653/v1/D19-5801} {{MRQA} 2019 shared
  task: Evaluating generalization in reading comprehension}.
\newblock In \emph{Proceedings of the 2nd Workshop on Machine Reading for
  Question Answering}, pages 1--13, Hong Kong, China. Association for
  Computational Linguistics.

\bibitem[{Fisch et~al.(2019{\natexlab{b}})Fisch, Talmor, Jia, Seo, Choi, and
  Chen}]{fisch2019mrqa}
Adam Fisch, Alon Talmor, Robin Jia, Minjoon Seo, Eunsol Choi, and Danqi Chen.
  2019{\natexlab{b}}.
\newblock Mrqa 2019 shared task: Evaluating generalization in reading
  comprehension.
\newblock In \emph{EMNLP 2019 MRQA Workshop}, page~1.

\bibitem[{Ganin et~al.(2017)Ganin, Ustinova, Ajakan, Germain, Larochelle,
  Laviolette, Marchand, and Lempitsky}]{Ganin2017}
Yaroslav Ganin, Evgeniya Ustinova, Hana Ajakan, Pascal Germain, Hugo
  Larochelle, Fran{\c{c}}ois Laviolette, Mario Marchand, and Victor Lempitsky.
  2017.
\newblock \href {https://doi.org/10.1007/978-3-319-58347-1_10}
  {\emph{Domain-Adversarial Training of Neural Networks}}, pages 189--209.
  Springer International Publishing, Cham.

\bibitem[{Golub et~al.(2017)Golub, Huang, He, and Deng}]{golub-etal-2017-two}
David Golub, Po-Sen Huang, Xiaodong He, and Li~Deng. 2017.
\newblock \href {https://doi.org/10.18653/v1/D17-1087} {Two-stage synthesis
  networks for transfer learning in machine comprehension}.
\newblock In \emph{Proceedings of the 2017 Conference on Empirical Methods in
  Natural Language Processing}, pages 835--844, Copenhagen, Denmark.
  Association for Computational Linguistics.

\bibitem[{Goodfellow et~al.(2014)Goodfellow, Pouget-Abadie, Mirza, Xu,
  Warde-Farley, Ozair, Courville, and Bengio}]{NIPS2014_5ca3e9b1}
Ian Goodfellow, Jean Pouget-Abadie, Mehdi Mirza, Bing Xu, David Warde-Farley,
  Sherjil Ozair, Aaron Courville, and Yoshua Bengio. 2014.
\newblock \href
  {https://proceedings.neurips.cc/paper/2014/file/5ca3e9b122f61f8f06494c97b1afccf3-Paper.pdf}
  {Generative adversarial nets}.
\newblock In \emph{Advances in Neural Information Processing Systems},
  volume~27. Curran Associates, Inc.

\bibitem[{Gretton et~al.(2006)Gretton, Borgwardt, Rasch, Sch\"{o}lkopf, and
  Smola}]{NIPS2006_e9fb2eda}
Arthur Gretton, Karsten Borgwardt, Malte Rasch, Bernhard Sch\"{o}lkopf, and
  Alex Smola. 2006.
\newblock \href
  {https://proceedings.neurips.cc/paper/2006/file/e9fb2eda3d9c55a0d89c98d6c54b5b3e-Paper.pdf}
  {A kernel method for the two-sample-problem}.
\newblock In \emph{Advances in Neural Information Processing Systems},
  volume~19. MIT Press.

\bibitem[{He et~al.(2020)He, Fan, Wu, Xie, and Girshick}]{9157636}
Kaiming He, Haoqi Fan, Yuxin Wu, Saining Xie, and Ross Girshick. 2020.
\newblock \href {https://doi.org/10.1109/CVPR42600.2020.00975} {Momentum
  contrast for unsupervised visual representation learning}.
\newblock In \emph{2020 IEEE/CVF Conference on Computer Vision and Pattern
  Recognition (CVPR)}, pages 9726--9735.

\bibitem[{Joshi et~al.(2017)Joshi, Choi, Weld, and
  Zettlemoyer}]{joshi-etal-2017-triviaqa}
Mandar Joshi, Eunsol Choi, Daniel Weld, and Luke Zettlemoyer. 2017.
\newblock \href {https://doi.org/10.18653/v1/P17-1147} {{T}rivia{QA}: A large
  scale distantly supervised challenge dataset for reading comprehension}.
\newblock In \emph{Proceedings of the 55th Annual Meeting of the Association
  for Computational Linguistics (Volume 1: Long Papers)}, pages 1601--1611,
  Vancouver, Canada. Association for Computational Linguistics.

\bibitem[{Kamath et~al.(2020)Kamath, Jia, and
  Liang}]{kamath-etal-2020-selective}
Amita Kamath, Robin Jia, and Percy Liang. 2020.
\newblock \href {https://doi.org/10.18653/v1/2020.acl-main.503} {Selective
  question answering under domain shift}.
\newblock In \emph{Proceedings of the 58th Annual Meeting of the Association
  for Computational Linguistics}, pages 5684--5696, Online. Association for
  Computational Linguistics.

\bibitem[{Kang et~al.(2019)Kang, Jiang, Yang, and Hauptmann}]{8954037}
G.~Kang, L.~Jiang, Y.~Yang, and A.~G. Hauptmann. 2019.
\newblock \href {https://doi.org/10.1109/CVPR.2019.00503} {Contrastive
  adaptation network for unsupervised domain adaptation}.
\newblock In \emph{2019 IEEE/CVF Conference on Computer Vision and Pattern
  Recognition (CVPR)}, pages 4888--4897, Los Alamitos, CA, USA. IEEE Computer
  Society.

\bibitem[{Keklik(2018)}]{keklik2018automatic}
Onur Keklik. 2018.
\newblock \emph{Automatic question generation using natural language processing
  techniques}.
\newblock Ph.D. thesis, Izmir Institute of Technology (Turkey).

\bibitem[{Kolouri et~al.(2019)Kolouri, Nadjahi, Simsekli, Badeau, and
  Rohde}]{kolouri2019generalized}
Soheil Kolouri, Kimia Nadjahi, Umut Simsekli, Roland Badeau, and Gustavo Rohde.
  2019.
\newblock Generalized sliced wasserstein distances.
\newblock \emph{Advances in neural information processing systems}, 32.

\bibitem[{Kratzwald et~al.(2019)Kratzwald, Eigenmann, and
  Feuerriegel}]{kratzwald-etal-2019-rankqa}
Bernhard Kratzwald, Anna Eigenmann, and Stefan Feuerriegel. 2019.
\newblock \href {https://doi.org/10.18653/v1/P19-1611} {{R}ank{QA}: Neural
  question answering with answer re-ranking}.
\newblock In \emph{Proceedings of the 57th Annual Meeting of the Association
  for Computational Linguistics}, pages 6076--6085, Florence, Italy.
  Association for Computational Linguistics.

\bibitem[{Kratzwald et~al.(2020)Kratzwald, Feuerriegel, and
  Sun}]{kratzwald-etal-2020-learning}
Bernhard Kratzwald, Stefan Feuerriegel, and Huan Sun. 2020.
\newblock \href {https://doi.org/10.18653/v1/2020.emnlp-main.246} {Learning a
  {C}ost-{E}ffective {A}nnotation {P}olicy for {Q}uestion {A}nswering}.
\newblock In \emph{Proceedings of the 2020 Conference on Empirical Methods in
  Natural Language Processing (EMNLP)}, pages 3051--3062, Online. Association
  for Computational Linguistics.

\bibitem[{Kwiatkowski et~al.(2019)Kwiatkowski, Palomaki, Redfield, Collins,
  Parikh, Alberti, Epstein, Polosukhin, Devlin, Lee, Toutanova, Jones, Kelcey,
  Chang, Dai, Uszkoreit, Le, and Petrov}]{10.1162/tacl_a_00276}
Tom Kwiatkowski, Jennimaria Palomaki, Olivia Redfield, Michael Collins, Ankur
  Parikh, Chris Alberti, Danielle Epstein, Illia Polosukhin, Jacob Devlin,
  Kenton Lee, Kristina Toutanova, Llion Jones, Matthew Kelcey, Ming-Wei Chang,
  Andrew~M. Dai, Jakob Uszkoreit, Quoc Le, and Slav Petrov. 2019.
\newblock \href {https://doi.org/10.1162/tacl_a_00276} {{Natural Questions: A
  Benchmark for Question Answering Research}}.
\newblock \emph{Transactions of the Association for Computational Linguistics},
  7:453--466.

\bibitem[{Lee et~al.(2019{\natexlab{a}})Lee, Batra, Baig, and
  Ulbricht}]{sliced-wasserstein-discrepancy-for-unsupervised-domain-adaptation}
Chen-Yu Lee, Tanmay Batra, Mohammad~Haris Baig, and Daniel Ulbricht.
  2019{\natexlab{a}}.
\newblock \href {https://arxiv.org/pdf/1903.04064} {Sliced wasserstein
  discrepancy for unsupervised domain adaptation}.

\bibitem[{Lee et~al.(2020)Lee, Lee, Jeong, Kim, and
  Hwang}]{lee-etal-2020-generating}
Dong~Bok Lee, Seanie Lee, Woo~Tae Jeong, Donghwan Kim, and Sung~Ju Hwang. 2020.
\newblock \href {https://doi.org/10.18653/v1/2020.acl-main.20} {Generating
  diverse and consistent {QA} pairs from contexts with information-maximizing
  hierarchical conditional {VAE}s}.
\newblock In \emph{Proceedings of the 58th Annual Meeting of the Association
  for Computational Linguistics}, pages 208--224, Online. Association for
  Computational Linguistics.

\bibitem[{Lee et~al.(2019{\natexlab{b}})Lee, Kim, and
  Park}]{lee-etal-2019-domain}
Seanie Lee, Donggyu Kim, and Jangwon Park. 2019{\natexlab{b}}.
\newblock \href {https://doi.org/10.18653/v1/D19-5826} {Domain-agnostic
  question-answering with adversarial training}.
\newblock In \emph{Proceedings of the 2nd Workshop on Machine Reading for
  Question Answering}, pages 196--202, Hong Kong, China. Association for
  Computational Linguistics.

\bibitem[{Long et~al.(2015)Long, Cao, Wang, and
  Jordan}]{10.5555/3045118.3045130}
Mingsheng Long, Yue Cao, Jianmin Wang, and Michael~I. Jordan. 2015.
\newblock Learning transferable features with deep adaptation networks.
\newblock In \emph{Proceedings of the 32nd International Conference on
  International Conference on Machine Learning - Volume 37}, ICML'15, page
  97–105. JMLR.org.

\bibitem[{Long et~al.(2018)Long, CAO, Wang, and Jordan}]{NEURIPS2018_ab88b157}
Mingsheng Long, ZHANGJIE CAO, Jianmin Wang, and Michael~I Jordan. 2018.
\newblock \href
  {https://proceedings.neurips.cc/paper/2018/file/ab88b15733f543179858600245108dd8-Paper.pdf}
  {Conditional adversarial domain adaptation}.
\newblock In \emph{Advances in Neural Information Processing Systems},
  volume~31. Curran Associates, Inc.

\bibitem[{Long et~al.(2017)Long, Zhu, Wang, and
  Jordan}]{10.5555/3305890.3305909}
Mingsheng Long, Han Zhu, Jianmin Wang, and Michael~I. Jordan. 2017.
\newblock Deep transfer learning with joint adaptation networks.
\newblock In \emph{Proceedings of the 34th International Conference on Machine
  Learning - Volume 70}, ICML'17, page 2208–2217. JMLR.org.

\bibitem[{McCann et~al.(2018)McCann, Keskar, Xiong, and
  Socher}]{DBLP:journals/corr/abs-1806-08730}
Bryan McCann, Nitish~Shirish Keskar, Caiming Xiong, and Richard Socher. 2018.
\newblock \href {http://arxiv.org/abs/1806.08730} {The natural language
  decathlon: Multitask learning as question answering}.
\newblock \emph{CoRR}, abs/1806.08730.

\bibitem[{Miller et~al.(2020)Miller, Krauth, Recht, and
  Schmidt}]{10.5555/3524938.3525579}
John Miller, Karl Krauth, Benjamin Recht, and Ludwig Schmidt. 2020.
\newblock The effect of natural distribution shift on question answering
  models.
\newblock In \emph{Proceedings of the 37th International Conference on Machine
  Learning}, ICML'20. JMLR.org.

\bibitem[{Pei et~al.(2018)Pei, Cao, Long, and Wang}]{10.5555/3504035.3504517}
Zhongyi Pei, Zhangjie Cao, Mingsheng Long, and Jianmin Wang. 2018.
\newblock Multi-adversarial domain adaptation.
\newblock In \emph{Proceedings of the Thirty-Second AAAI Conference on
  Artificial Intelligence and Thirtieth Innovative Applications of Artificial
  Intelligence Conference and Eighth AAAI Symposium on Educational Advances in
  Artificial Intelligence}, AAAI'18/IAAI'18/EAAI'18. AAAI Press.

\bibitem[{Peters et~al.(2018)Peters, Neumann, Iyyer, Gardner, Clark, Lee, and
  Zettlemoyer}]{peters-etal-2018-deep}
Matthew~E. Peters, Mark Neumann, Mohit Iyyer, Matt Gardner, Christopher Clark,
  Kenton Lee, and Luke Zettlemoyer. 2018.
\newblock \href {https://doi.org/10.18653/v1/N18-1202} {Deep contextualized
  word representations}.
\newblock In \emph{Proceedings of the 2018 Conference of the North {A}merican
  Chapter of the Association for Computational Linguistics: Human Language
  Technologies, Volume 1 (Long Papers)}, pages 2227--2237, New Orleans,
  Louisiana. Association for Computational Linguistics.

\bibitem[{Phang et~al.(2018)Phang, F{\'{e}}vry, and
  Bowman}]{DBLP:journals/corr/abs-1811-01088}
Jason Phang, Thibault F{\'{e}}vry, and Samuel~R. Bowman. 2018.
\newblock \href {http://arxiv.org/abs/1811.01088} {Sentence encoders on stilts:
  Supplementary training on intermediate labeled-data tasks}.
\newblock \emph{CoRR}, abs/1811.01088.

\bibitem[{Rajpurkar et~al.(2016)Rajpurkar, Zhang, Lopyrev, and
  Liang}]{rajpurkar-etal-2016-squad}
Pranav Rajpurkar, Jian Zhang, Konstantin Lopyrev, and Percy Liang. 2016.
\newblock \href {https://doi.org/10.18653/v1/D16-1264} {{SQ}u{AD}: 100,000+
  questions for machine comprehension of text}.
\newblock In \emph{Proceedings of the 2016 Conference on Empirical Methods in
  Natural Language Processing}, pages 2383--2392, Austin, Texas. Association
  for Computational Linguistics.

\bibitem[{Saito et~al.(2018)Saito, Watanabe, Ushiku, and Harada}]{8578490}
K.~Saito, K.~Watanabe, Y.~Ushiku, and T.~Harada. 2018.
\newblock \href {https://doi.org/10.1109/CVPR.2018.00392} {Maximum classifier
  discrepancy for unsupervised domain adaptation}.
\newblock In \emph{2018 IEEE/CVF Conference on Computer Vision and Pattern
  Recognition (CVPR)}, pages 3723--3732, Los Alamitos, CA, USA. IEEE Computer
  Society.

\bibitem[{Seo et~al.(2016)Seo, Kembhavi, Farhadi, and
  Hajishirzi}]{DBLP:journals/corr/SeoKFH16}
Min~Joon Seo, Aniruddha Kembhavi, Ali Farhadi, and Hannaneh Hajishirzi. 2016.
\newblock \href {http://arxiv.org/abs/1611.01603} {Bidirectional attention flow
  for machine comprehension}.
\newblock \emph{CoRR}, abs/1611.01603.

\bibitem[{Shakeri et~al.(2020)Shakeri, Nogueira~dos Santos, Zhu, Ng, Nan, Wang,
  Nallapati, and Xiang}]{shakeri-etal-2020-end}
Siamak Shakeri, Cicero Nogueira~dos Santos, Henghui Zhu, Patrick Ng, Feng Nan,
  Zhiguo Wang, Ramesh Nallapati, and Bing Xiang. 2020.
\newblock \href {https://doi.org/10.18653/v1/2020.emnlp-main.439} {End-to-end
  synthetic data generation for domain adaptation of question answering
  systems}.
\newblock In \emph{Proceedings of the 2020 Conference on Empirical Methods in
  Natural Language Processing (EMNLP)}, pages 5445--5460, Online. Association
  for Computational Linguistics.

\bibitem[{Shen et~al.(2018)Shen, Qu, Zhang, and Yu}]{10.5555/3504035.3504532}
Jian Shen, Yanru Qu, Weinan Zhang, and Yong Yu. 2018.
\newblock Wasserstein distance guided representation learning for domain
  adaptation.
\newblock In \emph{Proceedings of the Thirty-Second AAAI Conference on
  Artificial Intelligence and Thirtieth Innovative Applications of Artificial
  Intelligence Conference and Eighth AAAI Symposium on Educational Advances in
  Artificial Intelligence}, AAAI'18/IAAI'18/EAAI'18. AAAI Press.

\bibitem[{Shen et~al.(2022)Shen, Jones, Kumar, Xie, Haochen, Ma, and
  Liang}]{pmlr-v162-shen22d}
Kendrick Shen, Robbie~M Jones, Ananya Kumar, Sang~Michael Xie, Jeff~Z. Haochen,
  Tengyu Ma, and Percy Liang. 2022.
\newblock \href {https://proceedings.mlr.press/v162/shen22d.html} {Connect, not
  collapse: Explaining contrastive learning for unsupervised domain
  adaptation}.
\newblock In \emph{Proceedings of the 39th International Conference on Machine
  Learning}, volume 162 of \emph{Proceedings of Machine Learning Research},
  pages 19847--19878. PMLR.

\bibitem[{Sun et~al.(2018)Sun, Liu, Lyu, He, Ma, and
  Wang}]{sun-etal-2018-answer}
Xingwu Sun, Jing Liu, Yajuan Lyu, Wei He, Yanjun Ma, and Shi Wang. 2018.
\newblock \href {https://doi.org/10.18653/v1/D18-1427} {Answer-focused and
  position-aware neural question generation}.
\newblock In \emph{Proceedings of the 2018 Conference on Empirical Methods in
  Natural Language Processing}, pages 3930--3939, Brussels, Belgium.
  Association for Computational Linguistics.

\bibitem[{Talmor and Berant(2019)}]{talmor2019multiqa}
Alon Talmor and Jonathan Berant. 2019.
\newblock Multiqa: An empirical investigation of generalization and transfer in
  reading comprehension.
\newblock In \emph{Proceedings of the 57th Annual Meeting of the Association
  for Computational Linguistics}, pages 4911--4921.

\bibitem[{Tang et~al.(2017)Tang, Duan, Qin, and
  Zhou}]{DBLP:journals/corr/TangDQZ17}
Duyu Tang, Nan Duan, Tao Qin, and Ming Zhou. 2017.
\newblock \href {http://arxiv.org/abs/1706.02027} {Question answering and
  question generation as dual tasks}.
\newblock \emph{CoRR}, abs/1706.02027.

\bibitem[{Tang et~al.(2018)Tang, Duan, Yan, Zhang, Sun, Liu, Lv, and
  Zhou}]{tang-etal-2018-learning}
Duyu Tang, Nan Duan, Zhao Yan, Zhirui Zhang, Yibo Sun, Shujie Liu, Yuanhua Lv,
  and Ming Zhou. 2018.
\newblock \href {https://doi.org/10.18653/v1/N18-1141} {Learning to collaborate
  for question answering and asking}.
\newblock In \emph{Proceedings of the 2018 Conference of the North {A}merican
  Chapter of the Association for Computational Linguistics: Human Language
  Technologies, Volume 1 (Long Papers)}, pages 1564--1574, New Orleans,
  Louisiana. Association for Computational Linguistics.

\bibitem[{Thota and Leontidis(2021)}]{DBLP:journals/corr/abs-2103-15566}
Mamatha Thota and Georgios Leontidis. 2021.
\newblock \href {http://arxiv.org/abs/2103.15566} {Contrastive domain
  adaptation}.
\newblock \emph{CoRR}, abs/2103.15566.

\bibitem[{Trischler et~al.(2016)Trischler, Wang, Yuan, Harris, Sordoni,
  Bachman, and Suleman}]{https://doi.org/10.48550/arxiv.1611.09830}
Adam Trischler, Tong Wang, Xingdi Yuan, Justin Harris, Alessandro Sordoni,
  Philip Bachman, and Kaheer Suleman. 2016.
\newblock \href {https://doi.org/10.48550/ARXIV.1611.09830} {Newsqa: A machine
  comprehension dataset}.

\bibitem[{Tzeng et~al.(2017)Tzeng, Hoffman, Saenko, and Darrell}]{8099799}
E.~Tzeng, J.~Hoffman, K.~Saenko, and T.~Darrell. 2017.
\newblock \href {https://doi.org/10.1109/CVPR.2017.316} {Adversarial
  discriminative domain adaptation}.
\newblock In \emph{2017 IEEE Conference on Computer Vision and Pattern
  Recognition (CVPR)}, pages 2962--2971, Los Alamitos, CA, USA. IEEE Computer
  Society.

\bibitem[{Vaswani et~al.(2017)Vaswani, Shazeer, Parmar, Uszkoreit, Jones,
  Gomez, Kaiser, and Polosukhin}]{vaswani2017attention}
Ashish Vaswani, Noam Shazeer, Niki Parmar, Jakob Uszkoreit, Llion Jones,
  Aidan~N Gomez, {\L}ukasz Kaiser, and Illia Polosukhin. 2017.
\newblock Attention is all you need.
\newblock \emph{Advances in neural information processing systems}, 30.

\bibitem[{Wang et~al.(2018)Wang, Singh, Michael, Hill, Levy, and
  Bowman}]{wang-etal-2018-glue}
Alex Wang, Amanpreet Singh, Julian Michael, Felix Hill, Omer Levy, and Samuel
  Bowman. 2018.
\newblock \href {https://doi.org/10.18653/v1/W18-5446} {{GLUE}: A multi-task
  benchmark and analysis platform for natural language understanding}.
\newblock In \emph{Proceedings of the 2018 {EMNLP} Workshop {B}lackbox{NLP}:
  Analyzing and Interpreting Neural Networks for {NLP}}, pages 353--355,
  Brussels, Belgium. Association for Computational Linguistics.

\bibitem[{Wang et~al.(2021)Wang, Wu, Weng, Chen, Qi, and
  Jiang}]{DBLP:journals/corr/abs-2106-05528}
Rui Wang, Zuxuan Wu, Zejia Weng, Jingjing Chen, Guo{-}Jun Qi, and Yu{-}Gang
  Jiang. 2021.
\newblock \href {http://arxiv.org/abs/2106.05528} {Cross-domain contrastive
  learning for unsupervised domain adaptation}.
\newblock \emph{CoRR}, abs/2106.05528.

\bibitem[{Xie et~al.(2018)Xie, Zheng, Chen, and Chen}]{pmlr-v80-xie18c}
Shaoan Xie, Zibin Zheng, Liang Chen, and Chuan Chen. 2018.
\newblock \href {https://proceedings.mlr.press/v80/xie18c.html} {Learning
  semantic representations for unsupervised domain adaptation}.
\newblock In \emph{Proceedings of the 35th International Conference on Machine
  Learning}, volume~80 of \emph{Proceedings of Machine Learning Research},
  pages 5423--5432. PMLR.

\bibitem[{Xu et~al.(2019)Xu, Liu, Shen, Liu, and Gao}]{xu-etal-2019-multi}
Yichong Xu, Xiaodong Liu, Yelong Shen, Jingjing Liu, and Jianfeng Gao. 2019.
\newblock \href {https://doi.org/10.18653/v1/N19-1271} {Multi-task learning
  with sample re-weighting for machine reading comprehension}.
\newblock In \emph{Proceedings of the 2019 Conference of the North {A}merican
  Chapter of the Association for Computational Linguistics: Human Language
  Technologies, Volume 1 (Long and Short Papers)}, pages 2644--2655,
  Minneapolis, Minnesota. Association for Computational Linguistics.

\bibitem[{Yang et~al.(2020)Yang, Xia, Ding, and
  Ding}]{DBLP:journals/corr/abs-2002-04869}
Guanglei Yang, Haifeng Xia, Mingli Ding, and Zhengming Ding. 2020.
\newblock \href {http://arxiv.org/abs/2002.04869} {Bi-directional generation
  for unsupervised domain adaptation}.
\newblock \emph{CoRR}, abs/2002.04869.

\bibitem[{Yang et~al.(2018)Yang, Qi, Zhang, Bengio, Cohen, Salakhutdinov, and
  Manning}]{yang-etal-2018-hotpotqa}
Zhilin Yang, Peng Qi, Saizheng Zhang, Yoshua Bengio, William Cohen, Ruslan
  Salakhutdinov, and Christopher~D. Manning. 2018.
\newblock \href {https://doi.org/10.18653/v1/D18-1259} {{H}otpot{QA}: A dataset
  for diverse, explainable multi-hop question answering}.
\newblock In \emph{Proceedings of the 2018 Conference on Empirical Methods in
  Natural Language Processing}, pages 2369--2380, Brussels, Belgium.
  Association for Computational Linguistics.

\bibitem[{Yu et~al.(2020)Yu, Xu, Xu, Pang, Gao, Wang, and
  Wen}]{yu-etal-2020-wasserstein}
Weijie Yu, Chen Xu, Jun Xu, Liang Pang, Xiaopeng Gao, Xiaozhao Wang, and
  Ji-Rong Wen. 2020.
\newblock \href {https://doi.org/10.18653/v1/2020.emnlp-main.239}
  {{W}asserstein distance regularized sequence representation for text matching
  in asymmetrical domains}.
\newblock In \emph{Proceedings of the 2020 Conference on Empirical Methods in
  Natural Language Processing (EMNLP)}, pages 2985--2994, Online. Association
  for Computational Linguistics.

\bibitem[{Yue et~al.(2022{\natexlab{a}})Yue, Yao, and
  Sun}]{yue-etal-2022-synthetic}
Xiang Yue, Ziyu Yao, and Huan Sun. 2022{\natexlab{a}}.
\newblock \href {https://doi.org/10.18653/v1/2022.acl-long.95} {Synthetic
  question value estimation for domain adaptation of question answering}.
\newblock In \emph{Proceedings of the 60th Annual Meeting of the Association
  for Computational Linguistics (Volume 1: Long Papers)}, pages 1340--1351,
  Dublin, Ireland. Association for Computational Linguistics.

\bibitem[{Yue et~al.(2021)Yue, Kratzwald, and
  Feuerriegel}]{yue-etal-2021-contrastive}
Zhenrui Yue, Bernhard Kratzwald, and Stefan Feuerriegel. 2021.
\newblock \href {https://doi.org/10.18653/v1/2021.emnlp-main.754} {Contrastive
  domain adaptation for question answering using limited text corpora}.
\newblock In \emph{Proceedings of the 2021 Conference on Empirical Methods in
  Natural Language Processing}, pages 9575--9593, Online and Punta Cana,
  Dominican Republic. Association for Computational Linguistics.

\bibitem[{Yue et~al.(2022{\natexlab{b}})Yue, Zeng, Kou, Shang, and
  Wang}]{yue-etal-2022-domain}
Zhenrui Yue, Huimin Zeng, Ziyi Kou, Lanyu Shang, and Dong Wang.
  2022{\natexlab{b}}.
\newblock \href {https://aclanthology.org/2022.coling-1.153} {Domain adaptation
  for question answering via question classification}.
\newblock In \emph{Proceedings of the 29th International Conference on
  Computational Linguistics}, pages 1776--1790, Gyeongju, Republic of Korea.
  International Committee on Computational Linguistics.

\bibitem[{Yue et~al.(2022{\natexlab{c}})Yue, Zeng, Kratzwald, Feuerriegel, and
  Wang}]{yue2022qa}
Zhenrui Yue, Huimin Zeng, Bernhard Kratzwald, Stefan Feuerriegel, and Dong
  Wang. 2022{\natexlab{c}}.
\newblock Qa domain adaptation using hidden space augmentation and
  self-supervised contrastive adaptation.
\newblock \emph{arXiv preprint arXiv:2210.10861}.

\bibitem[{Zellinger et~al.(2017)Zellinger, Grubinger, Lughofer,
  Natschl{\"{a}}ger, and Saminger{-}Platz}]{DBLP:conf/iclr/ZellingerGLNS17}
Werner Zellinger, Thomas Grubinger, Edwin Lughofer, Thomas Natschl{\"{a}}ger,
  and Susanne Saminger{-}Platz. 2017.
\newblock \href {https://openreview.net/forum?id=SkB-\_mcel} {Central moment
  discrepancy {(CMD)} for domain-invariant representation learning}.
\newblock In \emph{5th International Conference on Learning Representations,
  {ICLR} 2017, Toulon, France, April 24-26, 2017, Conference Track
  Proceedings}. OpenReview.net.

\bibitem[{Zeng et~al.(2022)Zeng, Yue, Kou, Shang, Zhang, and
  Wang}]{https://doi.org/10.48550/arxiv.2210.03250}
Huimin Zeng, Zhenrui Yue, Ziyi Kou, Lanyu Shang, Yang Zhang, and Dong Wang.
  2022.
\newblock \href {https://doi.org/10.48550/ARXIV.2210.03250} {Unsupervised
  domain adaptation for covid-19 information service with contrastive
  adversarial domain mixup}.

\bibitem[{Zeng et~al.()Zeng, Yue, Zhang, Kou, Shang, and Wang}]{zengattacking}
Huimin Zeng, Zhenrui Yue, Yang Zhang, Ziyi Kou, Lanyu Shang, and Dong Wang.
\newblock On attacking out-domain uncertainty estimation in deep neural
  networks.

\end{thebibliography}
\bibliographystyle{acl_natbib}

\appendix

\section{Target Domain Dataset Details}
\label{sec:appendix}
The dataset details we have considered for target domain is given as follows:
\begin{itemize}
    \item \textbf{NaturalQuestions}: A real world QA dataset with questions that are actual user questions, and contexts as Wikipedia articles, which may or may not contain the answers \cite{10.1162/tacl_a_00276}
    \item \textbf{HotpotQA}: A reasoning based QA dataset with multi hop questions and supporting facts \cite{yang-etal-2018-hotpotqa}
    \item \textbf{SearchQA}: QA dataset where context built by crawling through Google Search. however, this is based on existing QA pairs to which the context is extended \cite{DBLP:journals/corr/DunnSHGCC17}
    \item \textbf{TriviaQA}: A reasoning based QA dataset containing evidences for questions asked \cite{joshi-etal-2017-triviaqa}
    \item \textbf{NewsQA}: QA dataset with news as contexts and questions with answers not from simple matching and entailment \cite{https://doi.org/10.48550/arxiv.1611.09830}
\end{itemize}

\end{document}